\documentclass[11pt,a4paper]{article}
\usepackage[hyperref]{acl2021}
\usepackage{times}
\usepackage{latexsym}

\usepackage{microtype}

\usepackage{bm}
\usepackage{booktabs} 
\usepackage{multirow,multicol}
\usepackage{subfigure}
\usepackage{kotex}
\usepackage{mfirstuc,tabulary}
\usepackage{graphicx}
\usepackage{amsmath}
\usepackage{xspace}
\usepackage{paralist}
\usepackage{multirow}
\usepackage{amsmath}
\usepackage{amsfonts}
\usepackage{pifont}
\usepackage{amssymb}
\usepackage{mathrsfs}
\usepackage{makecell}
\usepackage{enumitem}
\usepackage{algorithm,algpseudocode}
\usepackage{mdwlist}
\usepackage{CJK}
\usepackage{arydshln}
\usepackage{paralist}

\newcommand{\method}{LaSS\xspace}

\aclfinalcopy % Uncomment this line for the final submission
\title{Learning Language Specific Sub-network for Multilingual Machine Translation}
\author{Zehui Lin\footnotemark[1],
        Liwei Wu\thanks{\ \ Equal contribution.},
        Mingxuan Wang,
        Lei Li \\
    
    ByteDance AI Lab \\
    \textit{\{linzehui,wuliwei.000,wangmingxuan.89,lileilab\}@bytedance.com}
        
        }

\date{}

\begin{document}
\maketitle
\begin{abstract}

Multilingual neural machine translation aims at learning a single translation model for multiple languages.
These jointly trained models often suffer from performance degradation on rich-resource language pairs.
We attribute this degeneration to parameter interference. 
% That is,  parameters in a sharing space are not able to take responsibility for the representation of various languages. 
In this paper, we propose \method to jointly train a single unified multilingual MT model. 
\method learns  \textbf{La}nguage \textbf{S}pecific \textbf{S}ub-network (\method) for each language pair to counter parameter interference.  
% \method learns to find a sparse sub-network for each language pair. 
% The learned sub-networks are sparse and updated by corresponding language pairs during the training stage.
% As such, each parameter is responsible for a smaller language group. 
Comprehensive experiments on IWSLT and WMT datasets with various Transformer architectures show that 
\method obtains gains on 36 language pairs by up to 1.2 BLEU. 
Besides, \method shows its strong generalization performance at easy adaptation to new language pairs and zero-shot translation. \method boosts zero-shot translation with an average of 8.3 BLEU on 30 language pairs. Codes and trained models are available at \url{https://github.com/NLP-Playground/LaSS}.

% \begin{inparaenum}[a)]
% \item \method obtains gains on 36 language pairs by up to 1.2 BLEU over the strong baseline due to the alleviation of parameter interference. 
% \item \method can easily adapt to new languages without dramatic drop on existing languages.  
% \item \method obtains substantial gains on zero-shot translation by up to 26.5 BLEU\footnote{All codes, data and models will be released after the review.}.
% \end{inparaenum}
% Our further study shows the generalization and extensibility of \method.

%In this paper, we investigate the following question for MNMT: can we develop a universal compact MNMT model to serve all language pairs and reduce the negative interference for high-resource language pairs? 

%Different from the recent adapter approach, we propose \method, a multilingual CompAct and Sparse Translation model. \method alleviates the interference problem on neuron-level by automatically finding a sparse sub-network from the full network for each language pair. Specific neurons are only updated by related language pairs at training phase, and are used by related language pairs at inference phase. 
%By separating language-specific components from shared ones, \method can obtain consistent performance gains compared to single multilingual model.
\end{abstract}

\section{Introduction}

Neural machine translation (NMT) has been very successful for bilingual machine translation~\cite{DBLP:journals/corr/BahdanauCB14,DBLP:conf/nips/VaswaniSPUJGKP17,DBLP:journals/corr/WuSCLNMKCGMKSJL16,DBLP:journals/corr/abs-1803-05567,DBLP:conf/aaai/SuWXLHZ18,DBLP:conf/emnlp/Wang19}. 
Recent research has demonstrated the efficacy of multilingual NMT, which supports translation from multiple source languages into multiple target languages with a single model~\cite{johnson-etal-2017-googles,DBLP:conf/naacl/AharoniJF19,zhang-2020-improving-massive,DBLP:journals/corr/abs-2010-11125,DBLP:conf/acl/SiddhantBCFCKAW20}. 
Multilingual NMT enjoys the advantage of deployment. Further, the parameter sharing of multilingual NMT encourages transfer learning of different languages. 
An extreme case is zero-shot translation, where direct translation between a language pair never seen in training is possible~\cite{johnson-etal-2017-googles}. 

\begin{figure}[!t]
\centering
% \small
\subfigure[Full network]{
  \includegraphics[width=.40\linewidth]{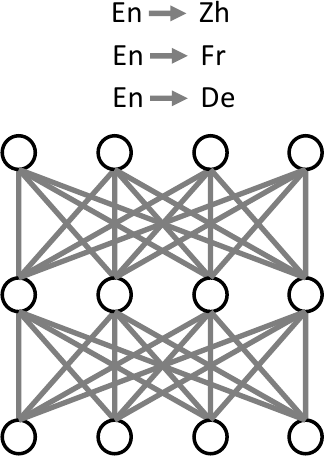}
  \label{fig:illustration2-fully}
  }
  \hspace{.2in}
  \subfigure[\method]{
  \includegraphics[width=.38\linewidth]{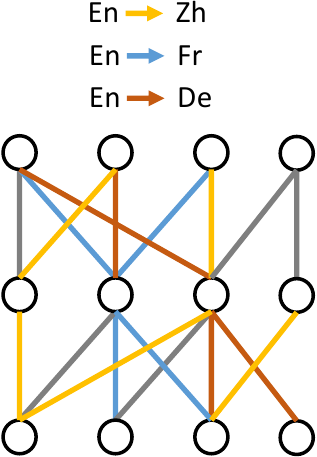}  
   \label{fig:illustration1-lass}
  }
\caption{Illustration of a full network and language-specific ones (\method). \textcolor[rgb]{0.498039216,0.498039216,0.498039216}{\textbf{---}} represents shared weights. \textcolor[rgb]{1,0.752941176,0}{\textbf{---}}, \textcolor[rgb]{0.356862745,0.607843137,0.831372549}{\textbf{---}} and \textcolor[rgb]{0.764705882,0.352941176,0.066666667}{\textbf{---}} represents weights for En\textcolor[rgb]{1,0.752941176,0}{\textbf{$\rightarrow$}}Zh, En\textcolor[rgb]{0.356862745,0.607843137,0.831372549}{$\rightarrow$}Fr and En\textcolor[rgb]{0.764705882,0.352941176,0.066666667}{$\rightarrow$}De, respectively. Compared to the full multilingual model, each \method learned model has language universal and language specific weights.}
\label{fig:illustration}
\end{figure}

While very promising, several challenges remain in multilingual NMT.  
The most challenging one is related to the insufficient model capacity. 
Since multiple languages are accommodated in a single model, the modeling capacity of NMT model has to be split for different translation directions~\cite{DBLP:conf/naacl/AharoniJF19}. 
Therefore, multilingual NMT models often suffer from performance degradation compared with their corresponding bilingual baseline, especially for rich-resource translation directions. 
The simplistic way to alleviate the insufficient model capacity is to enlarge the model parameters~\cite{DBLP:conf/naacl/AharoniJF19,zhang-2020-improving-massive}. However, it is not parameter or computation efficient and needs larger multilingual training datasets to avoid over-fitting. 
An alternative solution is to design language-aware components, such as division of the hidden cells into shared and language-dependent ones~\cite{DBLP:conf/emnlp/WangZZXZ18}, adaptation layers~\cite{DBLP:conf/emnlp/BapnaF19, DBLP:conf/emnlp/PhilipBGB20}, language-aware layer normalization
and linear transformation~\cite{zhang-2020-improving-massive}, and latent layers~\cite{DBLP:conf/nips/LiSTK20}. 

In this work, we propose \method, a method to dynamically find and learn \textbf{La}nguage  \textbf{S}pecific \textbf{S}ub-network  for multilingual NMT. 
\method accommodates one sub-network for each language pair. Each sub-network has shared parameters with some other languages and, at the same time, preserves its language specific parameters. 
In this way, multilingual NMT can model language specific and language universal features for each language pair in one single model without interference.
Figure~\ref{fig:illustration} is the illustration of vanilla multilingual model and \method. 
Each language pair in \method has both language universal and language specific parameters. 
The network itself decides the sharing strategy. 

The advantages of our proposed method are 
\begin{itemize*}
\item %Compared to previous work~\cite{DBLP:conf/emnlp/BapnaF19,zhang-2020-improving-massive,li2020deep}, 
\method is parameter efficient, requiring no extra trainable parameters to model language specific features.
\item \method alleviates parameter interference, potentially improving the model capacity and boosting performance. 
\item \method shows its strong generalization performance at easy adaptation to new language pairs and zero-shot translation. \method can be easily extended to new language pairs without dramatic degradation of existing language pairs. Besides, \method can boost zero-shot translation by up to 26.5 BLEU.
\end{itemize*}
% Different from previous approach,  \method let the network itself to learn the parameter sharing strategy. 
% The motivation of our idea lies in two aspects. 
% First, it is appealing to train  language specific sub-networks which achieve the same or better test accuracy compared with the original NMT model.  ~\citet{frankle2018lottery} discovered that a dense neural network contains a subnetwork which can match the test accuracy of the original network. 
% Further,  accommodating language specific sub-network within a single model can potentially improve the model capacity and avoid parameter conflicts. 

% \begin{figure*}[htbp]
%     \centering
%     \includegraphics[width=0.9\textwidth]{Figures/illustration-crop.pdf}
%     \caption{}
%     \label{fig:illustration}
% \end{figure*}

\section{Related Work}

\paragraph{Multilingual Neural Machine Translation}
The standard multilingual NMT model uses a shared encoder and a shared decoder for different languages~\cite{johnson-etal-2017-googles}. There is a transfer-interference trade-off in this architecture~\cite{DBLP:journals/corr/abs-1907-05019}: boosting the performance of low resource languages or maintain the performance of high resource languages. To solve this trade-off, previous works assign some parts of the model to be language specific: Language specific decoders~\cite{dong2015multi}, Language specific encoders and decoders~\cite{DBLP:conf/naacl/FiratCB16, DBLP:conf/emnlp/LyuSYB20} and Language specific hidden states and embeds~\cite{DBLP:conf/emnlp/WangZZXZ18}. \citet{DBLP:conf/wmt/SachanN18} compares different sharing methods and finds different sharing methods have a great impact on performance. Recently, \citet{zhang2021share} analyze when and where language specific capacity matters. \citet{DBLP:conf/nips/LiSTK20} uses a binary conditional latent variable to decide which language each layer belongs to.

\paragraph{Model Pruning}
Our approach follows the standard pattern of model pruning: training, finding the sparse network and fine-tuning~\cite{DBLP:conf/iclr/FrankleC19, DBLP:conf/iclr/LiuSZHD19}. \citet{DBLP:conf/iclr/FrankleC19} and \citet{DBLP:conf/iclr/LiuSZHD19} highlight the importance of the sparse network architecture. \citet{zhu2018prune} proposed a method to  automatically adjust the sparse threshold. \citet{DBLP:conf/aaai/SunSLLYQH20} learns different sparse architecture for different tasks. \citet{DBLP:conf/icml/EvciGMCE20} iteratively redistribute the sparse network architecture by the gradient.

\section{Methodology}
\label{sec:method}

We describe \method method in this section. 
The goal is to learn a single unified model for many translation directions. 
Our overall idea is to find sub-networks corresponding to each language pair, and 
then only update the parameters of those sub-networks during the joint training. 

\subsection{Multilingual NMT}
A multilingual NMT model learns a mapping function $f$ from a sentence in one of many languages to another language. 
We adopt the multilingual Transformer (mTransformer) as the backbone network \citep{johnson-etal-2017-googles}.
mTransformer has the same encoder-decoder architecture with layers of multihead attention, residual connection, and layer normalization. 
In addition, it has two lanuage identifying tokens for the source and target. 
Define a multilingual dataset $\{\mathcal{D}_{s_i \rightarrow t_i}\}_{i=1}^N$ where $s_i$, $t_i$ represents the source and target language.
% Define $L=\left\{L_{1}, \ldots, L_{M}\right\}$ where $L$ is the set of languages involved in training.
%     $\mathcal{D}_{i, j}$ denotes the parallel dataset of $(L_i,L_j)$, and $\mathcal{P}$ denotes the set of parallel datasets $\{\mathcal{D}\}_{k=1}^{N}$, where $N$ is the number of bilingual language pairs.

We train an initial multilingual MT model with the following loss.
\begin{equation}
    \mathcal{L}=\sum_{i} \sum_{\langle\mathbf{x}, \mathbf{y}\rangle \sim \mathcal{D}_{s_i \rightarrow t_i} }  -\log P_{\theta}(\mathbf{y} \mid \mathbf{x})
    \label{eq:multi}
\end{equation}
where $\langle\mathbf{x},\mathbf{y}\rangle$ is a sentence pair from the language $s_i$ to $t_i$, and $\theta$ is the model parameter.

\subsection{Finding Language Specific Model Masks}

% However, intuitively, jointly training on all translation pairs will obtain an "average" model on all directions due to the parameter interference with each other.

% \begin{figure}[htb]
%     \centering
%     \includegraphics[width=0.3\textwidth]{Figures/avg-crop.pdf}
%     \caption{Jointly training on all translation pairs obtains an ``average" MT model, deviating from the bilingual models.}
%     \label{fig:avg}
% \end{figure}

Training a single model jointly on multiple language directions will lead to performance degradation for rich resource pairs~\citep{johnson-etal-2017-googles}. 
The single model will improve on low resource language pairs, but will reduce performance on pairs like English-German. 
Intuitively, jointly training on all translation pairs will obtain an ``average" model.
% , which deviates from the bilingual models (Figure~\ref{fig:avg}).
For rich resources, such averaging may hurt the performance 
since a multilingual MT model must distribute its modeling capacity for all translation directions. 
Based on this intuition, our idea is to find a sub-network of the original multilingual model. 
Such sub-network is specific to each language pair.
% resembling embedding all models into one single model. 
% Each sub-network has shared parameters with some other languages and at the same time preserves its language specific parameters. In this way, the model is able to model language universal and language specific features without interference.

% We start from a multilingual base model $\mathcal{E}$, with its parameters $\theta_\mathcal{E}$. The $\mathcal{E}$ is trained by Eq.~\eqref{eq:multi}. 
We start from a multilingual base model $\theta_0$. The $\theta_0$ is trained with Eq.~\eqref{eq:multi}.
% We then assign a language specific sub-network $\mathcal{E}^{s_i \rightarrow t_i}$ for language pair $s_i\rightarrow t_i$. 
% We employ a binary mask matrix to indicate the sub-network.
A sub-network is indicated by a binary mask vector $\mathbf{M}_{s_i\rightarrow t_i} \in \{0,1\}^{\left|\theta \right|}$ for language pair $s_i \rightarrow t_i$.
% Denote a matrix $\mathbf{M}_{s_i\rightarrow t_i} \in \{0,1\}^{|\mathcal{E}^{s_i \rightarrow t_i}|}$ as the mask of $s_i\rightarrow t_i$.
Each element being $1$ indicates to retain the weight and $0$ to abandon the weight.
Then the parameters associated with $s_i\rightarrow t_i$ is $\theta_{s_i\rightarrow t_i}=\{ \theta_0^j \mid \mathbf{M}^j_{s_i\rightarrow t_i} =1  \}$, where $j$ denotes the $j$th element in $\theta_0$.
% Then the parameters associated with $s_i\rightarrow t_i$ is  $\theta_{\mathcal{E}^{s_i \rightarrow t_i}} = \mathbf{M}_{s_i\rightarrow t_i} \odot \theta_\mathcal{E}$.
The parameters $\theta_{s_i \rightarrow t_i}$ are only responsible for the particular language $s_i$ and $t_i$. 
We intend to find such language specific sub-networks. 
Figure~\ref{fig:illustration} illustrates the original model and its language specific sub-networks.

% The parameters $\theta_{s_i \rightarrow t_i}$ are only updated by the corresponding language pairs $s_i\rightarrow t_i$ at training phase, and are used by it at inference phase.
% After obtaining sub-networks for each language pair, we jointly train the sub-networks in parallel until convergence. 
% Specifically, the language specific parameters $\theta_{\mathcal{E}^{i\rightarrow j}}$ is calcuated by $\mathbf{M}_{i \rightarrow j} \odot \theta_{\mathcal{E}}$ where $\mathbf{M}_{i\rightarrow j} \in \{0,1\}^{|\mathcal{E}^{i \rightarrow j}|}$ is a binary mask. For the direction $i\rightarrow j$, only the language specific parameters $\theta_{\mathcal{E}^{i\rightarrow j}}$ are updated at training phase and used at inference phase.

% Thus, in general, \method contains two phases.
% \begin{inparaenum}[a)]
% \item Search for masks for each language pair.
% \item Jointly train the sub-networks in parallel.
% \end{inparaenum}

% \paragraph{Search for masks for each language pair}

% Previous works have explored finding network (sub-)structure for certain tasks \cite{DBLP:conf/iclr/ZophL17, DBLP:conf/aaai/RealAHL19, DBLP:conf/eccv/MallyaDL18, DBLP:conf/emnlp/ZhaoLMJS20, DBLP:conf/aaai/SunSLLYQH20}.
% Ideally, sub-networks corresponding to similar language pairs\footnote{We define ``similarity" according to commonly used language family.}(like En$\rightarrow$Fr and En$\rightarrow$Es) should share larger portion of parameters than those corresponding to dissimilar language pairs(like En$\rightarrow$Fr and En$\rightarrow$Zh).

Given an initial model $\theta_0$, we adopt a simple method to find the language specific mask for each language pairs.
\begin{enumerate*}
    \item Start with a multilingual MT model $\theta_0$ jointly trained on $\{\mathcal{D}_{s_i \rightarrow t_i}\}_{i=1}^N$.
    \item For each language pair $s_i \rightarrow t_i$, fine-tuning $\theta_0$ on $\mathcal{D}_{s_i \rightarrow t_i}$. Intuitively, fine-tuning $\theta_0$ on specific language pair $s_i\rightarrow t_i$ will amplify the magnitude of the important weights for $s_i\rightarrow t_i$ and diminish the magnitude of the unimportant weights. 
    \item Rank the weights in fine-tuned model and prune the lowest $\alpha$ percent. The mask $\mathbf{M}_{s_i \rightarrow t_i}$ is obtained by setting the remaining indices of parameters to be 1. 
\end{enumerate*}

% We start with a vanilla ``average" multilingual MT model $\theta_0$.
% We then continue training (fine-tuning) $\theta_0$ on the each language pair $s_i\rightarrow t_i$ respectively for fixed steps to obtain a set of ``optimal" models $\theta^{*}_{s_i \rightarrow t_i}$ for each language pair.  
% Intuitively, fine-tuning $\theta_0$ on specific language pair $s_i\rightarrow t_i$ will amplify the magnitude of the important weights for $s_i\rightarrow t_i$ and diminish the magnitude of the unimportant weights. 
% After obtaining a set of "optimal" models, we can prune the unimportant weights by the magnitude for each language pair. We prune $\alpha$ percent with the lowest magnitudes from $\theta^*_{s_i \rightarrow t_i}$, generating $\mathbf{M}_{s_i \rightarrow t_i}$. 

% In this way, each language pairs has its own sub-network $\theta_{s_i\rightarrow t_i}=\{ \theta_0 \mid \mathbf{M}^j_{s_i\rightarrow t_i} =1  \}$.

% Algorithm \ref{alg:method} shows the process of searching for masks.
 
\subsection{Structure-aware Joint Training}

Once we get masks $\mathbf{M}_{s_i \rightarrow t_i}$ for all language pairs, we further continue to train $\theta_0$ with language-grouped batching and structure-aware updating. 

First, we create random batches of bilingual sentence pairs where each batch contains only samples from one pair. 
This is different from the plain joint multilingual training where each batch can contain fully random sentence pairs from all languages. 
Specifically, a batch $\mathcal{B}_{s_i \rightarrow t_i}$ is randomly drawn from the language-specific data $\mathcal{D}_{s_i \rightarrow t_i}$. 
Second, we evaluate the loss in Eq.~\ref{eq:multi} on the batch $\mathcal{B}_{s_i \rightarrow t_i}$. During the back-propagation step, we only update the parameters in $\theta_0$ belonging to the sub-network indicated by $\mathbf{M}_{s_i\rightarrow t_i}$. 
We iteratively update the parameters until convergence.

In this way, we still get a single final model $\theta^*$ that is able to translate all language directions. 

During the inference, this model $\theta^*$ and its masks $\mathbf{M}_{s_i\rightarrow t_i}, i=1,\dots,N$ are used together to make predictions.
For every given input sentence in language $s$ and a target language $t$, 
the forward inference step only uses the parameter $\theta^* \odot \mathbf{M}_{s\rightarrow t}$ to calculate model output.  
% \method contains four phase: 
% 1) Same as vanilla multilingual nmt, we first train a base "average" model $\mathcal{E}$ that can support multilingual translation.
% 2) We then continue training the "average" model $\mathcal{E}$ on the each language pairs $(L_i,L_j)$ for fixed steps to obtain an "optimal" model $\mathcal{E}_{i \rightarrow j}$. Intuitively, it will amplify the magnitude of the important weights for the specific language pairs and diminish the magnitude of the unimportant weights from  $\Theta$. 
% 3) After obtaining a set of "optimal" models, we can prune the unimportant weights by the magnitude. Specifically, let $ M_{i \rightarrow j} \in \{0,1\}^{\left|\Theta_{i \rightarrow j}\right|}$ be the binary mask of model $\Theta_{i \rightarrow j}$. Where $1$ denotes preserving the weight while $0$ represents pruning the weight. In this way, each language pairs has its own sub-network $\Theta \odot M_{i \rightarrow j.} $
% 4) 

\section{Experiment Settings}
\label{sec:exp}

\paragraph{Datasets and Evaluation}
The experiments are conducted on IWSLT and WMT benchmarks.
For IWSLT, we collect 8 English-centric language pairs from IWSLT2014, whose size ranges from 89k to 169k.
To simulate the scenarios of imbalanced datasets, we collect 18 language pairs ranging from low-resource (Gu, 11k) to rich-resource (Fr, 37m) from previous years' WMT.
The details of the datasets are listed in Appendix.
We apply byte pair encoding (BPE) \cite{sennrich2015neural} to preprocess multilingual sentences, resulting in a vocabulary size of 30k for IWSLT and 64k for WMT. 
%we add two artiﬁcial language tokens to indicate specific languages at both the source and the target side.
Besides, we apply over-sampling for IWSLT and WMT to balance the training data distribution with a temperature of $T=2$ and $T=5$ respectively.
%\cite{DBLP:journals/corr/abs-1907-05019}.
Similar to \citet{DBLP:conf/emnlp/LinPWQFZL20}, we divide the language pairs into 3 categories: low-resource ($<$1M), medium-resource ($>$1M and $<$10M) and rich resource ($>$10M).

We perform many-to-many multilingual translation throughout this paper,
and add special language tokens at  both the source and the target side. 
In all our experiments, we evaluate our model with commonly used standard testsets. For zero-shot, where standard testsets (for example, Fr$\rightarrow$Zh) of some language pairs are not available, we use OPUS-100 \cite{zhang-2020-improving-massive} testsets instead. 

We report tokenized BLEU, as well as win ratio (WR), informing the proportion of language pairs we outperform the baseline. In zero-shot translation, we also report translation-language accuracy\footnote{\url{https://github.com/Mimino666/langdetect}}, which is commonly used to measure the accuracy of translating into the right target language.

\paragraph{Model Settings}
Considering the diversity of dataset volume, we perform our experiments with variants of Transformer architecture. 
For IWSLT, we adopt a smaller Transformer (Transformer-small\footnote{Transformer-base with $d_{ff}=1024$ and $n_{head}=4$}~\cite{DBLP:conf/iclr/WuFBDA19}). 
For WMT, we adopt Transformer-base and Transformer-big\footnote{For details of the Transformer setting, please refer to \citet{DBLP:conf/nips/VaswaniSPUJGKP17}}. 
The pruning rate $\alpha$ of IWSLT and WMT is 0.7 and 0.3, respectively. For simplicity, we only report the highest BLEU from the best pruning rate and we also discuss the impact of different pruning rate on performance in Sec.\ref{sec:analysis}.
% For architecture,  we replace ReLU with GeLU and replace fixed positional embedding with learnable positional embedding.
In Sec.~\ref{sec:analysis} we discuss the relationship of performance and pruning rate.
For more training details please refer to Appendix.

\section{Experiment Results}
This section shows the efficacy and generalization of \method. Firstly, we show that \method obtains consistent performance gains on IWSLT and WMT datasets with different Transformer architecture variants.
Further, we show that \method can easily generalize to new language pairs without losing the accuracy for previous language pairs. 
Finally, we observe that \method can even improve zero-shot translation, obtaining performance gains by up to 26.5 BLEU.
%\subsection{Experiment Settings}

\subsection{Main Results}

\paragraph{Results on IWSLT}
We first show our results on IWSLT. As shown in Table \ref{tab:iwslt-result}, \method consistently outperforms the multilingual baseline on all language pairs, confirming that using \method to alleviate parameter interference can help boost performance.

% \begin{table*}[htb]
% % \center
% \begin{center}
% % \small
% % \scalebox{1.0}{
% \begin{tabular}{rcccccccccccc}
% \toprule

% Lang-Pairs& 
% \mf{En-Fa} &
% \mf{En-Pl} &
% \mf{En-Ar} &
% \mf{En-He} & 
% Avg

% \\

% Size &
% \smf{89K}  & 
% \smf{128k}  & 
% \smf{140K} & 
% \smf{144K} & 

% \\

% Direction & 
%  $\rightarrow$ &$\leftarrow$ &

%  $\rightarrow$ &$\leftarrow$ &

%  $\rightarrow$ &$\leftarrow$ &

%  $\rightarrow$ &$\leftarrow$ \\

% \midrule
% baseline &
% 13.9 & 19.9 & % fa
% \bf 14.3 & 18.5 &  % pl
% 14.6 & 27.1 & % ar
% 25.9 & 32.1 & % he
% -
% \\

%  \method &
% \bf 14.0 & \bf 21.8 &
%  14.2 & \bf 19.7 &
% \bf 15.5 & \bf 30.2 &
% \bf 26.4 & \bf 35.4 &
% -

% \\

%  $\Delta$ &
%  +0.1 & +1.9 &
%  -0.1 & +1.2  &
%  +0.9 & +3.1 &
%  +0.5 & + 3.3 &
%  -

% \\

% \midrule
% \midrule

% Lang-Pairs &
% \mf{En-Nl} &
% \mf{En-De} &
% \mf{En-It} &
% \mf{En-Es} &
% Avg
% \\

% Size &
% \smf{153K} & 
% \smf{160K} & 
% \smf{167K} & 
% \smf{169K}
% \\

%  Direction & 
%  $\rightarrow$ &$\leftarrow$ &

%  $\rightarrow$ &$\leftarrow$ &

%  $\rightarrow$ &$\leftarrow$ &

%  $\rightarrow$ &$\leftarrow$ \\

% \midrule
%  baseline &
% 30.9 & 30.8 & %nl
% 27.5 & 28.7 & %de
% 29.3 & 29.1 & %it
% 36.2 & 34.2 & %es
% -
% \\

%  \method &
% \bf 31.4 & \bf 34.5 & %de
% \bf 27.5 & \bf 32.0 &  %nl
% \bf 29.5 & \bf 32.2 & %it
% \bf 36.2 & \bf 38.4 & %es
% -

% \\

% $\Delta$ &
% +0.5 & +3.7 &
% +0.0 & +3.3  &
% +0.2 & +3.1 &
% +0.0 & +4.2 &
% -

% \\

% \bottomrule
% \end{tabular}
% % }
% \caption{IWSLT}
% \label{tab:iwslt-result}
% \end{center}
% \end{table*}

\begin{table}[tb]
% \center
\begin{center}
% \small
% \scalebox{1.0}{
\begin{tabular}{rcccc}
\toprule

Lang& 
Fa &
Pl &
Ar &
He

\\

Size &
89K  & 
128k  & 
140K & 
144K 

\\

% Direction & 
%  $\rightarrow$ &$\leftarrow$ &

%  $\rightarrow$ &$\leftarrow$ &

%  $\rightarrow$ &$\leftarrow$ &

%  $\rightarrow$ &$\leftarrow$ \\

\midrule
Baseline &
16.9 & % fa
16.4 &  % pl
20.9 & % ar
29  % he

\\

 \method &
\bf 17.9 &
\bf 17.0 &
\bf 22.9 &
\bf 30.9

\\

 $\Delta$ &
 +1.0 &
 +0.6  &
 +2.0 &
 +1.9  

\\

\midrule
\midrule

Lang &
Nl &
De &
It &
Es
\\

Size &
153K & 
160K & 
167K & 
169K
\\

% %  Direction & 
% %  $\rightarrow$ &$\leftarrow$ &

% %  $\rightarrow$ &$\leftarrow$ &

% %  $\rightarrow$ &$\leftarrow$ &

% %  $\rightarrow$ &$\leftarrow$ \\

\midrule
 Baseline &
30.9 & %nl
28.1 & %de
29.2 & %it
35.2  %es
\\

 \method &
\bf 33.0 & %de
\bf 29.8 &  %nl
\bf 30.9 & %it
\bf 37.3  %es

\\

$\Delta$ &
+2.1 &
+1.7  &
+1.7 &
+2.1

\\

\bottomrule
\end{tabular}
% }
\caption{Results on IWLST dataset. Baseline denotes the multilingual Transformer-small baseline model. \method consistently outperforms multilingual baseline on all language pairs. We report the average BLEU of En$\rightarrow$X and X$\rightarrow$En within one language. Both the baseline and \method have the same number of parameters.}
\label{tab:iwslt-result}
\end{center}
\end{table}

\paragraph{Results on WMT}
To further verify the generalization of \method, we also conduct experiments on WMT, where the dataset is more imbalanced across different language pairs. We adopt two different Transformer architecture variants, i.e., Transformer-base and Transformer-big.

As shown in Table \ref{tab:wmt-result}, \method obtains consistent gains over multilingual baseline on WMT for both Transformer-base and Transformer-big. For Transformer-base, \method achieves an average improvement of 1.2 BLEU on 36 language pairs over baseline, while for Transformer-big, \method obtains 0.6 BLEU improvement.

We observe that with the dataset scale of language pairs increasing, the improvements of BLEU and WR become larger, suggesting that the language pairs with large scale dataset benefit more from \method than language pairs of low resource.
This phenomenon is intuitive since rich resource dataset suffers more parameter interference than low resource dataset.
We also find that the BLEU and WR gains obtained in Transformer-base are larger than that in Transformer-large. 
We attribute it to the more severe parameter interference for smaller models.

For comparison, we also include the results of \method with randomly initialized masks. Not surprising, Random underperforms the baseline by a large margin, since Random intensifies rather than alleviates the parameter interference.

\begin{table*}[ht]
\begin{tabular}{llllllllll}
\toprule
\multirow{2}{*}{Arch Setting}                      & \multirow{2}{*}{Model}  & \multicolumn{2}{c}{Low} & \multicolumn{2}{c}{Medium} & \multicolumn{2}{c}{Rich} & \multicolumn{2}{c}{All} \\ 
 &  & BLEU & WR & BLEU & WR & BLEU & WR & BLEU & WR \\
\hline
\multirow{3}{*}{Transformer-base}
& Baseline   & 16.7  & - & 18.8  & -  & 25.3 & - & 20.4  &  \\ 

& Random  & -2.2 & 0.0  & -2.3 & 0.0  & -2.6  & 0.0   & -2.4 & 0.0 \\ 

& \method & \bf +0.7 & \bf 80.0 & \bf +1.3 & \bf 85.7  & \bf +1.7 & \bf 100.0  & \bf +1.2  & \bf88.9  \\ 
\hline
\multirow{3}{*}{Transformer-big}  
& Baseline  & 18.8  & -   & 22.2 & -  & 29.0  & -  & 23.5  & -   \\ 

& Random   & -1.3   & 0.0  & -1.8  & 0.0  & -1.5  & 0.0    & -1.6     & 0.0 
\\ 
& \method & \bf +0.1  & \bf 50.0 & \bf +0.7   & \bf 92.9   & \bf +0.8  & \bf 100.0   & \bf +0.6    & \bf 83.3       \\ 
\bottomrule
\end{tabular}

\caption{Average BLEU$\uparrow$ and Win Ratio (WR) of WMT dataset on Low ($<$1M), Medium (1M$\sim$10M) and Rich ($>$10M) resource dataset. Random denotes \method with random masks. \method obtains consistent gains for both Transformer-big and Transformer-base. }
\label{tab:wmt-result}
\end{table*}

\subsection{Generalization to New Language Pairs}

\method has shown its efficacy in the above section. 
A natural question arises that can \method adapt to a new language or language pair that it has not seen in training phase? In other words, can \method generalize to other language pairs?
% \method distributes different sub-network for different language pairs and alleviates the parameter interference. 
In this section, we show the generalization of \method in two settings.
We firstly show that \method can easily adapt to new \textbf{unseen} languages to match bilingual models with training for only a few hundred steps while keeping the performance of the existing language pairs hardly dropping.
Secondly, we show that \method can also boost performance in \textbf{zero-shot} translation scenario, obtaining performance gains by up to 26.5 BLEU.
% We aims to answer two questions: 1) Can \method adapt to new language pair with training for only a few hundred steps while keeping the performance of other language pairs unchanged? 2) To take a step further, can \method adapt to a new language pairs without training? 

The model is Transformer-big trained on WMT dataset. En$\leftrightarrow$Ar and En$\leftrightarrow$It are both unseen language pairs.

\subsubsection{Extensibility to New Languages}

Previous works have studied the easy and rapid adaptation to a new task or language pair \cite{DBLP:conf/emnlp/BapnaF19, DBLP:conf/nips/RebuffiBV17}. 
We show that \method can also easily adapt to new unseen languages without dramatic drop for other existing languages.
% todo, Difference: Multilingual base model, Multilingual baseline model
We distribute a new sub-network to each new language pair and train the sub-network with the specific language pair for fixed steps. In this way, the new language pair will only update the corresponding parameters and it can alleviate the interference and catastrophic forgetting~\cite{DBLP:journals/corr/KirkpatrickPRVD16} to other language pairs.

We verify the extensibility of \method on 4 language pairs.
For \method, as described in Sec.\ref{sec:method},  we first fine-tune the multilingual base model and prune to obtain the specific mask for the new language pair.
For both multilingual baseline and our method, we train on only the specific language pair for fixed steps.

\begin{figure}[!t] \centering
\subfigure[En$\rightarrow$It]{
  \includegraphics[width=.45\linewidth]{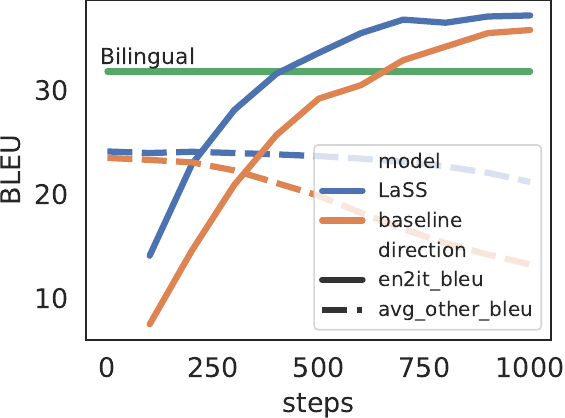}  
  \label{fig:en2it}
  }
\subfigure[It$\rightarrow$En]{
  \includegraphics[width=.45\linewidth]{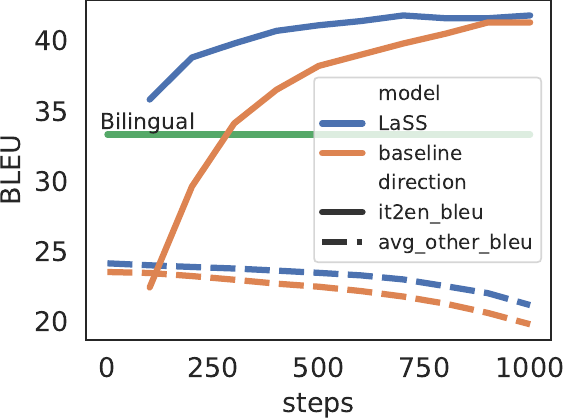}  
  \label{fig:it2en}
  }
\subfigure[En$\rightarrow$Ar]{
\includegraphics[width=.45\linewidth]{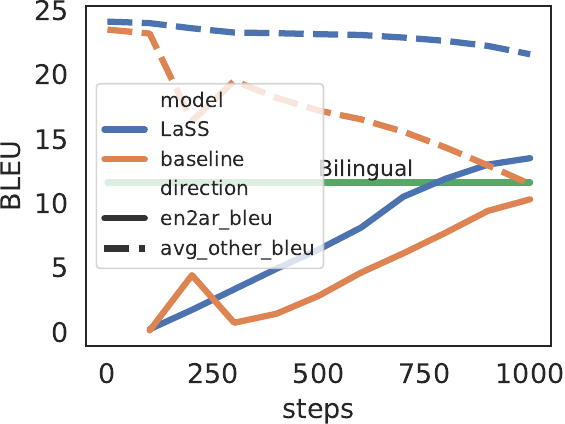}  
\label{fig:en2ar}
}
\subfigure[Ar$\rightarrow$En]{
  \includegraphics[width=.45\linewidth]{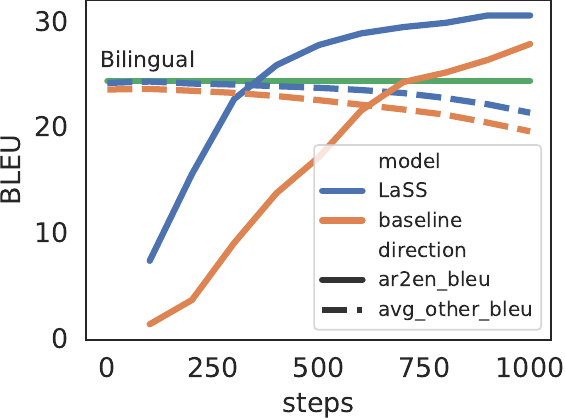}  
  \label{fig:ar2en}
  }
\caption{The trend of BLEU score of new extended language pairs and other existing language pairs along with the training steps on the specific language pair. Compared to multilingual baseline, \method reaches the bilingual performance with fewer steps and only little performance degradation on other existing language pairs.}
\label{fig:extensibility}
\end{figure}

Figure~\ref{fig:extensibility} shows the trend of BLEU score along with the training steps.
We observe that 1) \method consistently outperforms the multilingual baseline model along with the training steps. \method reaches the bilingual model performance with fewer steps. 2) Besides, the degradation of other language pairs is much smoother than the baseline. When reaching the bilingual baseline performance, \method hardly drops on other language pairs, while the multilingual baseline model dramatically drops by a large margin.

We attribute the easy adaptation for specific languages to the language specific sub-network. \method only updates the corresponding parameters, avoiding updating all parameters which will hurt the performance of other languages. Another benefit of updating corresponding parameters is its fast adaptation towards specific language pairs.

\subsubsection{Zero-shot}
Zero-shot translation is the translation between known languages that the model has never seen together at training time (e.g., Fr$\rightarrow$En and En$\rightarrow$Zh are both seen in training phase, while Fr$\rightarrow$Zh is not.). 
It is the ultimate goal of Multilingual NMT and has been a common indicator to measure the model capability \cite{johnson-etal-2017-googles, zhang-2020-improving-massive}.
One of the biggest challenges is the off-target issue~\cite{zhang-2020-improving-massive}, which means that the model translates into a wrong target language.

In previous experiments, we apply specific masks to their corresponding language pairs. As the training dataset is English-centric, non-English-centric masks are not available. We remedy it by merging two masks to create non-English-centric masks. For example, We create X$\rightarrow$Y mask by combining the encoder mask of X$\rightarrow$En and the decoder mask of En$\rightarrow$Y.  
We select 6 languages and evaluate zero-shot translation in language pairs between each other.

As shown in Table \ref{tab:zero-shot}, surprisingly, by directly applying X$\rightarrow$Y masks, \method obtains consistent gains over baselines in all language pairs for both BLEU and translation-language accuracy, indicating that the superiority of \method in learning to bridge between languages. 
It is worth noting that for Fr$\rightarrow$Zh, \method outperforms the baseline by 26.5 BLEU, reaching 32 BLEU.

\begin{table*}[ht]
\small
\renewcommand{\arraystretch}{1.2}
\setlength{\tabcolsep}{5pt}
\begin{tabular}{ll|lllllllllllll}
\toprule
\multicolumn{2}{l|}{\multirow{2}{*}{}}                   &                        & \multicolumn{12}{c}{Target Languages}                                                                                                                                                            \\
\multicolumn{2}{l|}{}                                    &                        & \multicolumn{2}{c}{Fr}         & \multicolumn{2}{c}{Cs}        & \multicolumn{2}{c}{De}        & \multicolumn{2}{c}{Es}         & \multicolumn{2}{c}{Ru}        & \multicolumn{2}{c}{Zh}         \\ 
\midrule
                                   &                     &                        & BLEU           & ACC           & BLEU          & ACC           & BLEU          & ACC           & BLEU           & ACC           & BLEU          & ACC           & BLEU           & ACC           \\
\multirow{18}{*}{\rotatebox[origin=c]{90}{Source Languages}} & \multirow{3}{*}{Fr} & baseline               & -              & -             & 2.0           & 1.7          & 2.9           & 3.1          & 6.4            & 15.1          & 1.5           & 4.4          & 5.5            & 4.9          \\
                                   &                     & \method & -              & -             & 5.4           & 32.6          & 7.5           & 35.9          & 23.0             & 77.7          & 4.6           & 24.7          & 32.0           & 31.3          \\
                                   &                     & $\Delta$               & -              & -    & \textbf{+3.4} & \textbf{+30.9} & \textbf{+4.6} & \textbf{+32.8} & \textbf{+16.6} & \textbf{+62.6} & \textbf{+3.1} & \textbf{20.3} & \textbf{+26.5} & \textbf{+26.4} \\ \cdashline{3-15} 
                                   & \multirow{3}{*}{Cs} & baseline               & 3.9            & 7.0          & -             & -             & 2.6           & 2.1          & 5.6            & 13.9          & 2.5           & 9.6          & 0.9            & 0.9          \\
                                   &                     & \method & 15.3           & 61.1          & -             & -             & 7.7           & 37.2          & 18.5           & 74.2          & 6.6           & 34.5          & 13.5           & 35.3          \\
                                   &                     & $\Delta$               & \textbf{+11.4} & \textbf{+54.1} & -             & -    & \textbf{+5.1} & \textbf{+35.1} & \textbf{+12.9} & \textbf{+60.3}  & \textbf{+4.1} & \textbf{+24.9} & \textbf{+12.6} & \textbf{+34.4} \\ \cdashline{3-15} 
                                   & \multirow{3}{*}{De} & baseline               & 6.3            & 18.8          & 2.6           & 5.7          & -             & -             & 5.6            & 14.0          & 2.2           & 8.6          & 5.7            & 19.6          \\
                                   &                     & \method & 17.9           & 70.3          & 7.4           & 40.5          & -             & -             & 19.4           & 75.1          & 6.1           & 33.2          & 16.1           & 41.6          \\
                                   &                     & $\Delta$               & \textbf{+11.6} & \textbf{+51.5} & \textbf{+4.8} & \textbf{+34.8} & -             & -    & \textbf{+13.8} & \textbf{+61.1} & \textbf{+3.9} & \textbf{+24.6} & \textbf{+10.4} & \textbf{+22.0} \\ \cdashline{3-15} 
                                   & \multirow{3}{*}{Es} & baseline               & 7.4            & 17.5          & 2.0             & 1.6          & 2.6           & 1.9          & -              & -             & 1.4           & 3.7         & 3.6            & 9.2          \\
                                   &                     & \method & 20.8           & 66.3          & 4.9           & 25.7          & 6.7           & 30.3          & -              & -             & 4.5           & 22.2          & 15.2           & 42.8          \\
                                   &                     & $\Delta$               & \textbf{+13.4} & \textbf{+48.8} & \textbf{+2.9} & \textbf{+24.1} & \textbf{+4.1} & \textbf{+28.4} & -              & -    & \textbf{+3.1} & \textbf{+18.5} & \textbf{+11.6} & \textbf{+33.6} \\ \cdashline{3-15} 
                                   & \multirow{3}{*}{Ru} & baseline               & 5.6            & 19.9          & 2.4           & 8.1          & 2.0             & 2.4          & 6.3            & 20.6          & -             & -             & 10.5           & 13.4          \\
                                   &                     & \method & 16.2           & 69.0          & 8.0             & 47.7          & 5.9           & 32.0          & 18.8           & 75.5          & -             & -             & 30.0             & 33.1          \\
                                   &                     & $\Delta$               & \textbf{+10.6} & \textbf{+49.1} & \textbf{+5.6} & \textbf{+39.6}  & \textbf{+3.9} & \textbf{+29.6}  & \textbf{+12.5} & \textbf{+54.9} & -             & -    & \textbf{+19.5} & \textbf{+19.7}  \\ \cdashline{3-15} 
                                   & \multirow{3}{*}{Zh} & baseline               & 5.6            & 4.0          & 0.3           & 1.0          & 1.1           & 1.6          & 0.8            & 2.1          & 4.8           & 5.6          & -              & -             \\
                                   &                     & \method & 18             & 53.2         & 1.7           & 22.9          & 1.2           & 7.1          & 3.8            & 28.0          & 7.2           & 27.6          & -              & -             \\
                                   &                     & $\Delta$               & \textbf{+12.4} & \textbf{+49.2} & \textbf{+1.4} & \textbf{+21.9} & \textbf{+0.1} & \textbf{+5.5} & \textbf{+3.0}    & \textbf{+25.9} & \textbf{+2.4} & \textbf{+22.0} & -              & -             \\ 
                                   
  \bottomrule
\end{tabular}
\caption{BLEU score and Translation-language Accuracy (ACC, in percentage) of zero-shot translation for multilingual baseline and \method. \method outperforms the multilingual baseline on both BLEU and ACC by a large margin for most language pairs. Low accuracy indicates severe off-target translation.}
\label{tab:zero-shot}
\end{table*}

We also sample a few translation examples from Fr$\rightarrow$Zh to analyze why \method can help boost zero-shot (More examples are listed in Appendix).

\begin{CJK}{UTF8}{gkai}
\begin{table}[!ht]
\small
% \rowcolors{2}{lightgray}{}
\begin{center}
\begin{tabular}{lp{5.5cm}}
\toprule
Src & La production annuelle d'acier était le symbole incontesté de la vigueur économique des nations. \\
Ref & 钢的年产量是国家经济实力的重要象征 \\
Baseline & Annual steel production was the undisputed symbol of nations' economic strength. \\
\method & 年度钢铁生产是各国经济活力的无可争辩的象征. \\

\midrule

Src & De l'avis de ma délégation donc, l'ONU devrait élargir ces activités de la façon suivante. \\
Ref & 因此,我国代表团认为,联合国现在应该以下述方式扩大这些活动。 \\
Baseline & 因此, in my delegation's view, the United Nations should expand these activities in the following manner. \\
\method & 因此,我国代表团认为,联合国应该扩大这些活动,如下. \\

% \midrule

% Src & Le domicile de la femme dépendait du lieu du mariage et de la résidence familiale. \\
% Ref & 妇女的住处取决于婚姻和家庭位置。 \\
% Baseline & The woman's place of residence depended on the place of marriage and family residence. \\
% \method & 妻子的住所取决于婚姻地点和家庭住所. \\

\bottomrule
\end{tabular}
\caption{Fr$\rightarrow$Zh Case Study. The multilingual baseline suffers from severe off-target issue, while \method greatly alleviates the issue.}
\label{tab:fr2zh-case}
\end{center}
\end{table}
\end{CJK}

As shown in Table \ref{tab:fr2zh-case} as well as translation-language accuracy in Table~\ref{tab:zero-shot}, we observe that the multilingual baseline has severe off-target issue.
As a counterpart, \method significantly alleviates the off-target issue, translating into the right target language. 
We attribute the success of ``on-target" in zero-shot to the language specific parameters as a strong signal, apart from language indicator, to the model to translate into the target language.

\section{Analysis and Discussion}
\label{sec:analysis}
\begin{figure}[!t] \centering
\subfigure[En$\rightarrow$X (x-axis and y-axis)]{
  \includegraphics[width=.3\linewidth]{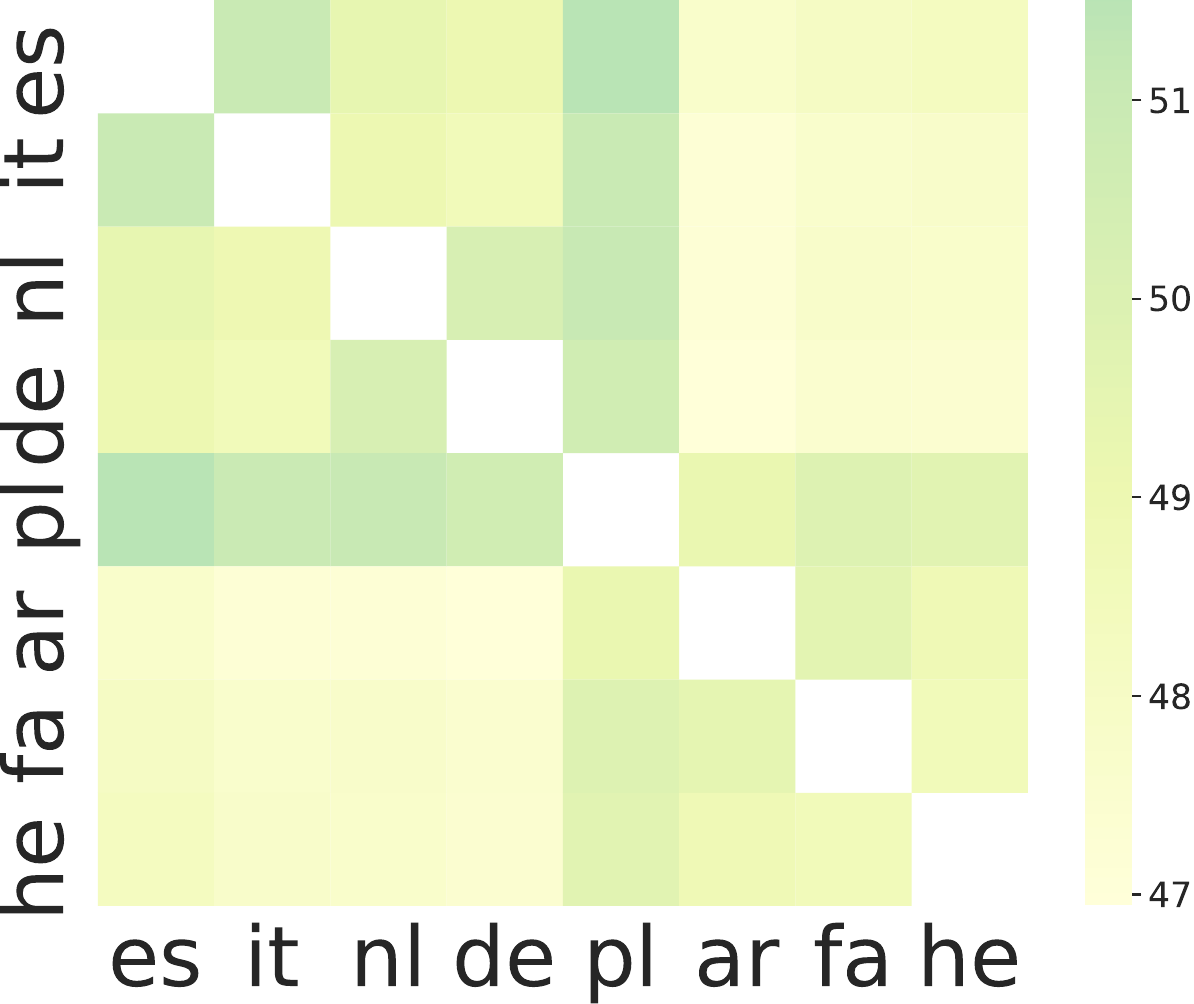}  
  \label{fig:en2x}
  }
\subfigure[X$\rightarrow$En (x-axis and y-axis)]{
  \includegraphics[width=.3\linewidth]{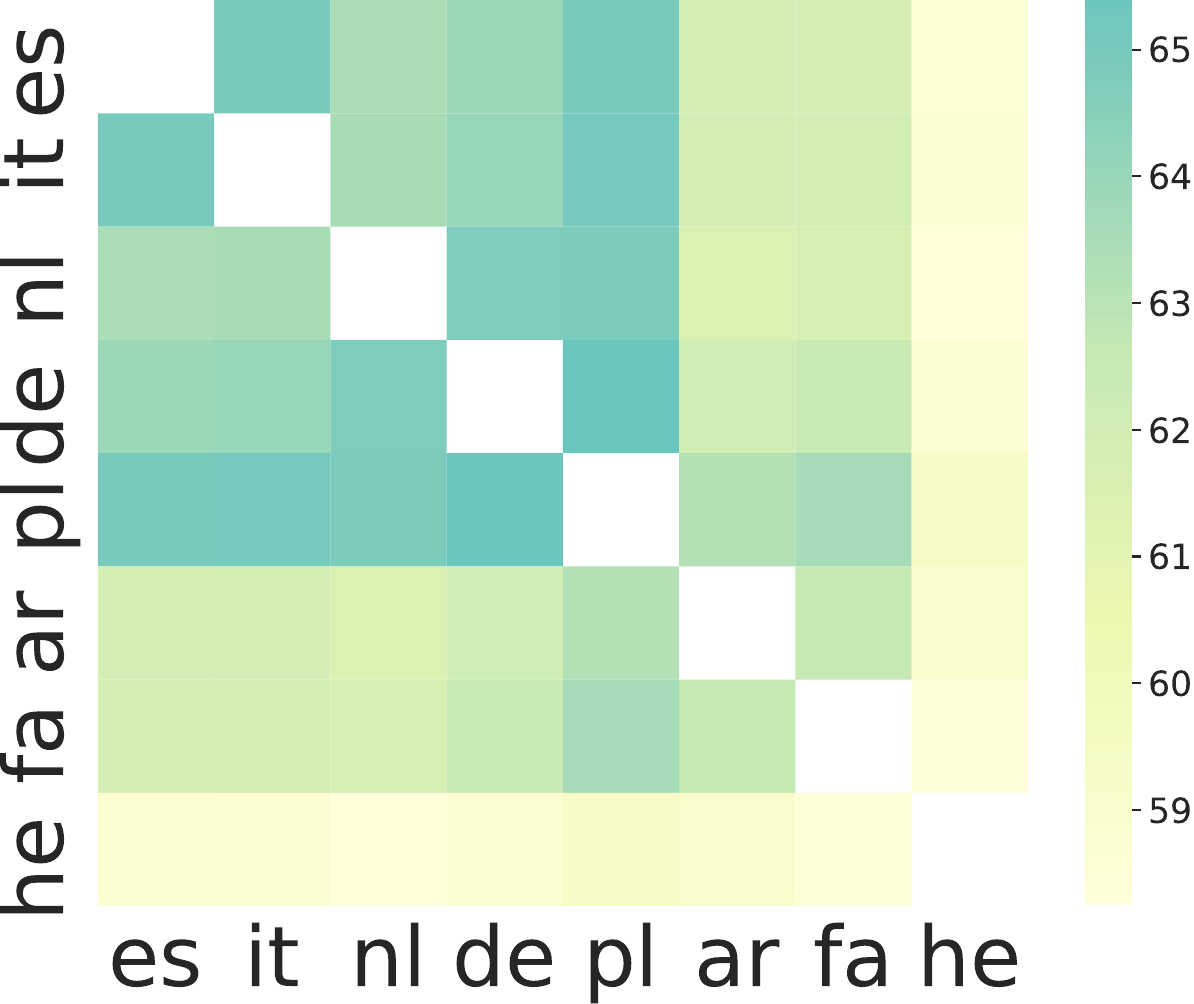}  
  \label{fig:x2en}
  }
  \subfigure[X$\rightarrow$En (x-axis) En$\rightarrow$X(y-axis) ]{
  \includegraphics[width=.3\linewidth]{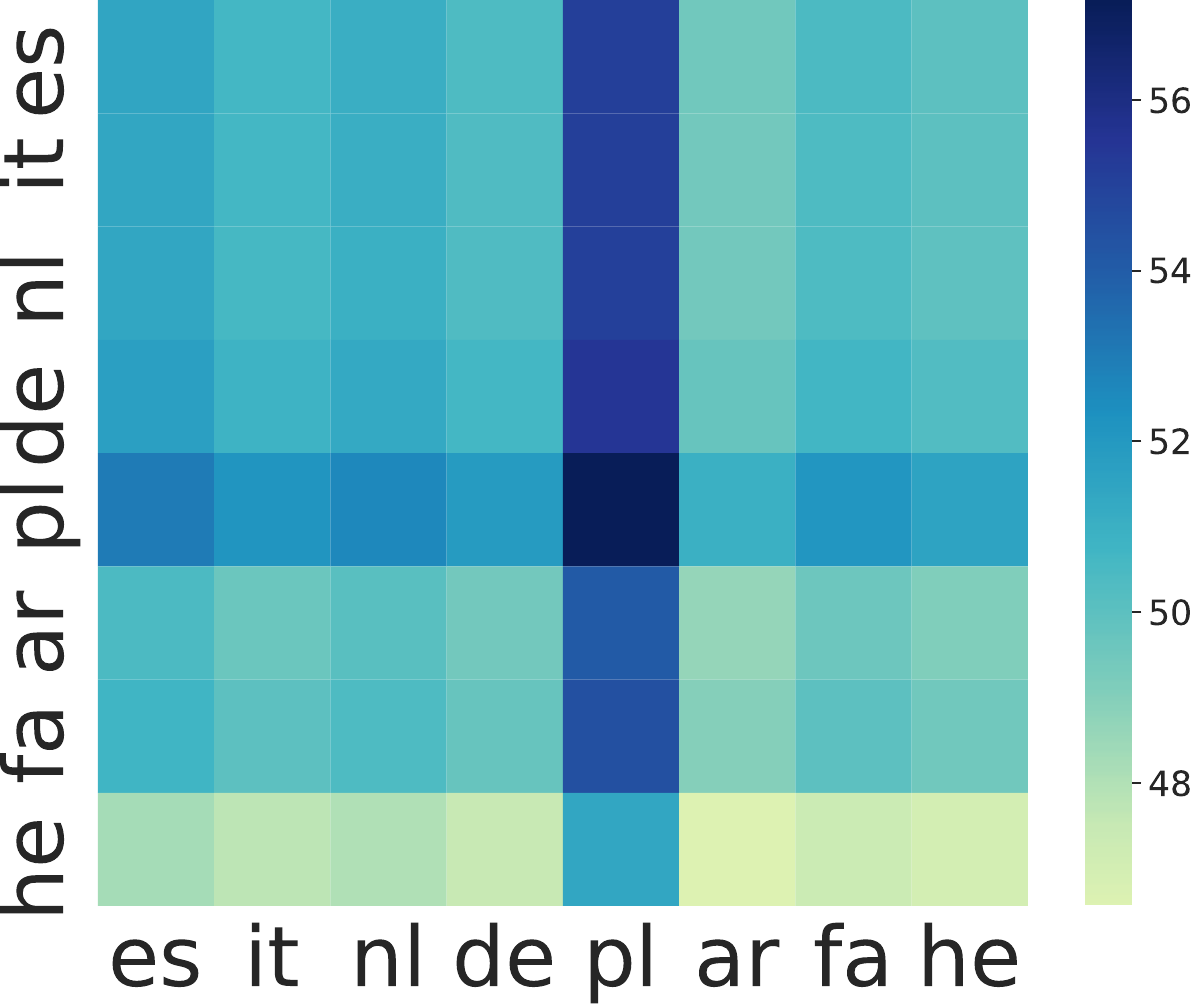}  
  \label{fig:x2en(x-axis)-en2x(y-axis)}
  }
\caption{Mask similarity for language pairs within En$\rightarrow$X (x-axis and y-axis), within X$\rightarrow$En (x-axis and y-axis) and between En$\rightarrow$X (x-axis) and X$\rightarrow$En (y-axis), respectively. The mask similarity is positively correlated to the language family similarity.}
\label{fig:mask-sim-whole}
\end{figure}

In this section, we conduct a set of analytic experiments to better understand the characteristics of language specific sub-network. 
We first measure the relationship between language specific sub-network as well as its capacity and language family. Secondly, we study how masks affect performance in zero-shot scenario. Lastly, we discuss the relationship between pruning rate $\alpha$ and performance.

\begin{figure}[!tbp]
    \centering
    \includegraphics[width=0.4\textwidth]{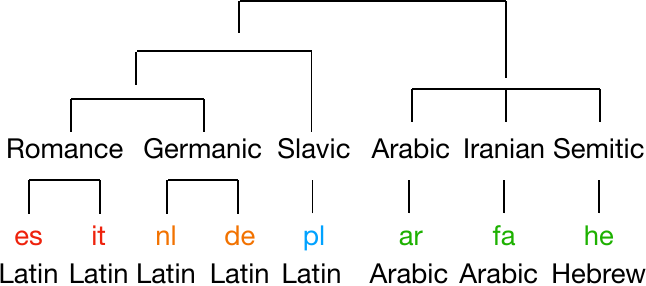}
    \caption{Language clustering of 8 languages in IWSLT, according to language family. Es(Spanish), It(Italian), De(Germany), Nl(Dutch) and Pl(Polish) are all European languages and written in Latin while Ar(Arabic), Fa(Farsi) and He(Hebrew) are similar languages.}
    \label{fig:lang-hie}
\end{figure}

We conduct our analytic experiments on IWSLT dataset. For readers not familiar with language family and clustering, Figure \ref{fig:lang-hie} is the hierarchical clustering according to language family. 

\subsection{Mask similarity v.s Language family}

Ideally, similar languages should share more parameters since they share more language characteristics.
Therefore, a natural question arises: Does the model automatically capture the relationship of language family defined by human?

We calculate the similarity of masks between language pairs to measure the sub-network relationship between language pairs. 
We define mask similarity as the number of $1$ where two masks share divided by the number of $1$ of the first mask:

\begin{equation}
    \text{Sim}(\mathbf{M}_1,\mathbf{M}_2)= \frac{\left \| \mathbf{M}_1 \cap  \mathbf{M}_2 \right \|_{0} }{\left \| \mathbf{M}_1 \right \|_{0}},
\end{equation}
where $\left \| \cdot  \right \|_{0}$ represent $L_0$ norm.
Mask similarity reflects the degree of sharing among different language pairs.
% Since the pruning rate of the two masks are the same, the equation is symmetric.

Figure \ref{fig:en2x} and \ref{fig:x2en} shows the mask similarity in En$\rightarrow $X and X$\rightarrow $En. We observe that, for both En$\rightarrow $X and X$\rightarrow $En, the mask similarity is positively correlated to the language family similarity. The color of grids in Figure is deeper between similar languages (for example, es and it) while more shallow between dissimilar languages (for example, es and he).

We also plot the similarity between En$\rightarrow $X and X$\rightarrow $En in Figure \ref{fig:x2en(x-axis)-en2x(y-axis)} . We observe that, unlike En$\rightarrow $X or X$\rightarrow $En, the mask similarity does not correspond to language family similarity. We suspect that the mask similarity is determined by \textbf{combination} of source and target languages. That means that En$\rightarrow$Nl does not necessarily share more parameters with Nl$\rightarrow$En than En$\rightarrow$De.

\subsection{Where language specific capacity matters?}

\begin{figure}[!t] \centering
\subfigure[Encoder]{
  \includegraphics[width=0.7\linewidth]{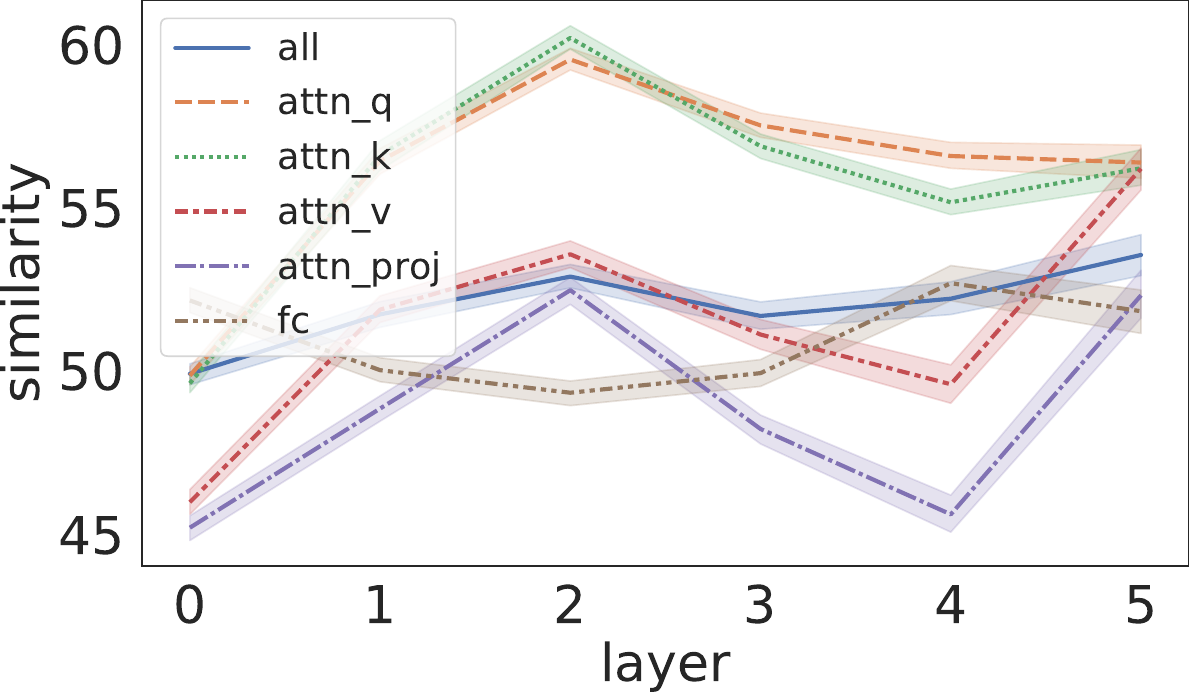}  
  \label{fig:component-encoder}
  }
\subfigure[Decoder]{
  \includegraphics[width=0.7\linewidth]{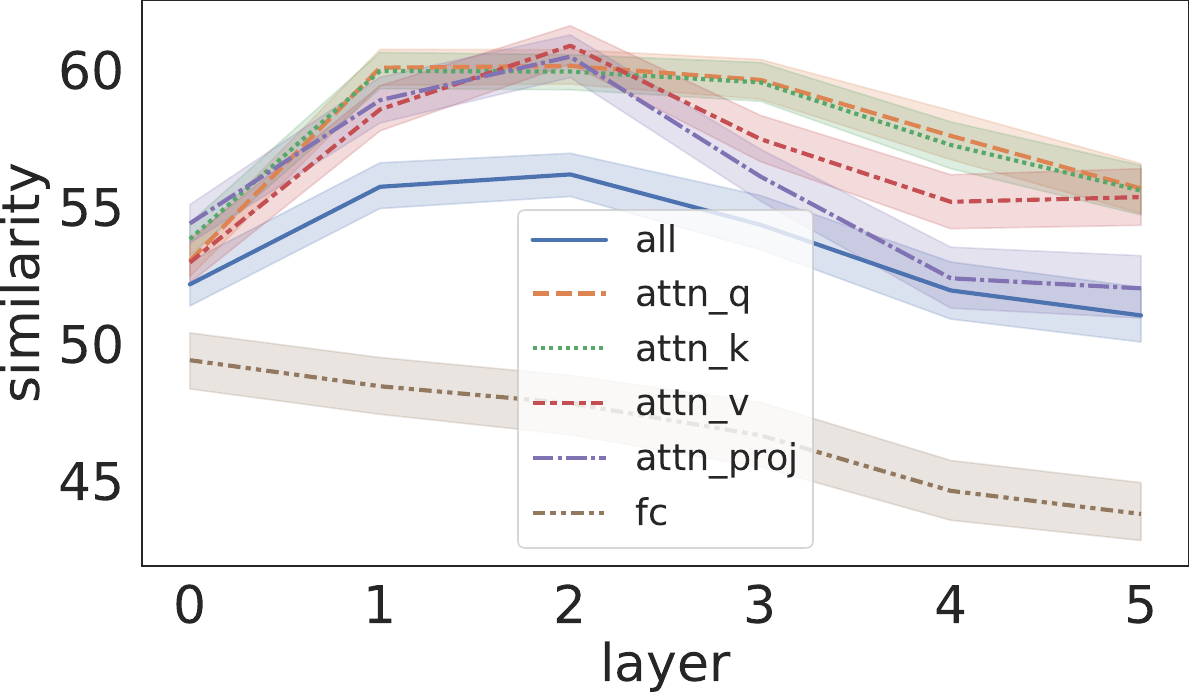}
  \label{fig:component-decoder}
  }
\caption{The mask similarity of different components (attention layer and feed-forward layer) on the encoder and decoder side along with the increase of layer. The model tends to distribute more language specific capacity on the top and bottom layers.}
\label{fig:mask-sim-component}
\end{figure}

% todo: plot similarity .. query, key, value on the attention sub-layer and fully-connected layer on the positional-wise feed-forward sub-layer. 
To take a step further, we study how model schedule language specific capacity across layers. 
Figure~\ref{fig:mask-sim-component} shows the similarity of different components on the encoder and decoder side along with the increase of layer. More concretely, we plot query, key, value on the attention sub-layer and fully-connected layer on the positional-wise feed-forward sub-layer.

We observe that 
\begin{inparaenum}[a)]
\item On both the encoder and decoder side, the model tends to distribute more language specific components on the top and bottom layers rather than the middle ones. This phenomenon is intuitive. The bottom layers deal more with embedding, which is language specific, while the top layers are near the output layer, which is also language specific.
\item For fully-connected layer, the model tends to distribute more language specific capacity on the middle layers for the encoder, while distribute more language specific capacity in the decoder for the top layers.
\end{inparaenum}

\subsection{How masks affect zero-shot?}
In Sec.\ref{sec:exp}, we show that simply applying X$\rightarrow$Y masks can boost zero-shot performance. 
We conduct experiments to analyze how masks affect zero-performance. Concretely, we take Fr$\rightarrow$Zh as an example, replacing the encoder or decoder mask with another language mask, respectively.

\begin{table}[!t]
\centering
\scalebox{0.85}{
\begin{tabular}{llllll}
\toprule
\multicolumn{6}{c}{\textbf{Fr $\rightarrow$X}}                     \\ 
\midrule
Fr            & Cs   & De   & Es   & Ru   & Zh            \\
-             & 12.3 & 13.8 & 7.1  & 18.6 & \textbf{32.0} \\ 
\midrule
\midrule
\multicolumn{6}{c}{\textbf{X $\rightarrow$ Zh}}                    \\ 
\midrule

Fr            & Cs   & De   & Es   & Ru   & Zh            \\
\textbf{32.0} & 30.5 & 29.6 & 30.9 & 29.6 & -             \\ 
\bottomrule
\end{tabular}}
\caption{Performance of applying Fr$\rightarrow$X or X$\rightarrow$Zh mask to Fr$\rightarrow$Zh testset. Replacing encoder mask causes only little performance drop, while replacing decoder mask causes dramatic performance drop.}
\label{tab:fr2zh-mask}
\end{table}

As shown in Table~\ref{tab:fr2zh-mask}, we observe that replacing the encoder mask with other languages causes only littler performance drop, while replacing the decoder mask causes dramatic performance drop. It suggests that the decoder mask is the key ingredient of performance improvement.

\subsection{About Sparsity}
\begin{figure}[!t] \centering
\subfigure[IWSLT]{
  \includegraphics[width=1.0\linewidth]{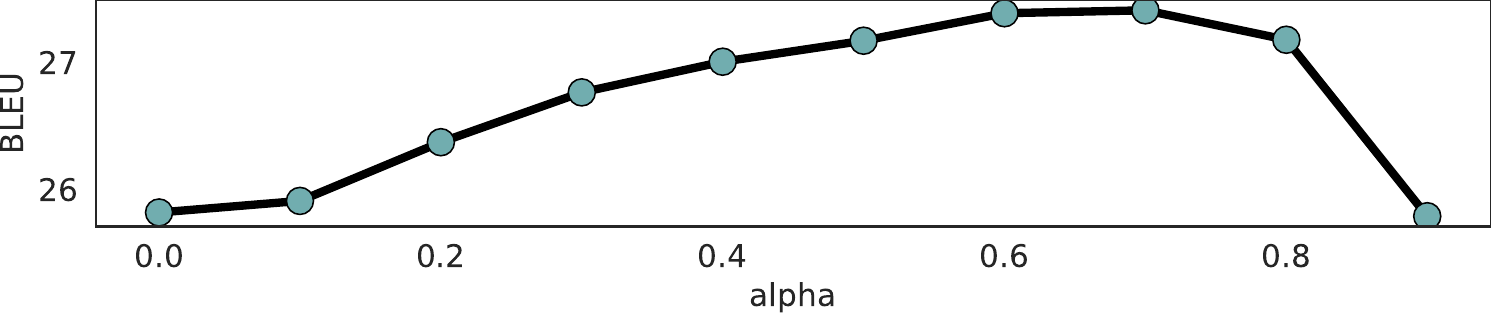}  
  \label{fig:alpha-iwslt}
  }
\subfigure[WMT]{
  \includegraphics[width=1.0\linewidth]{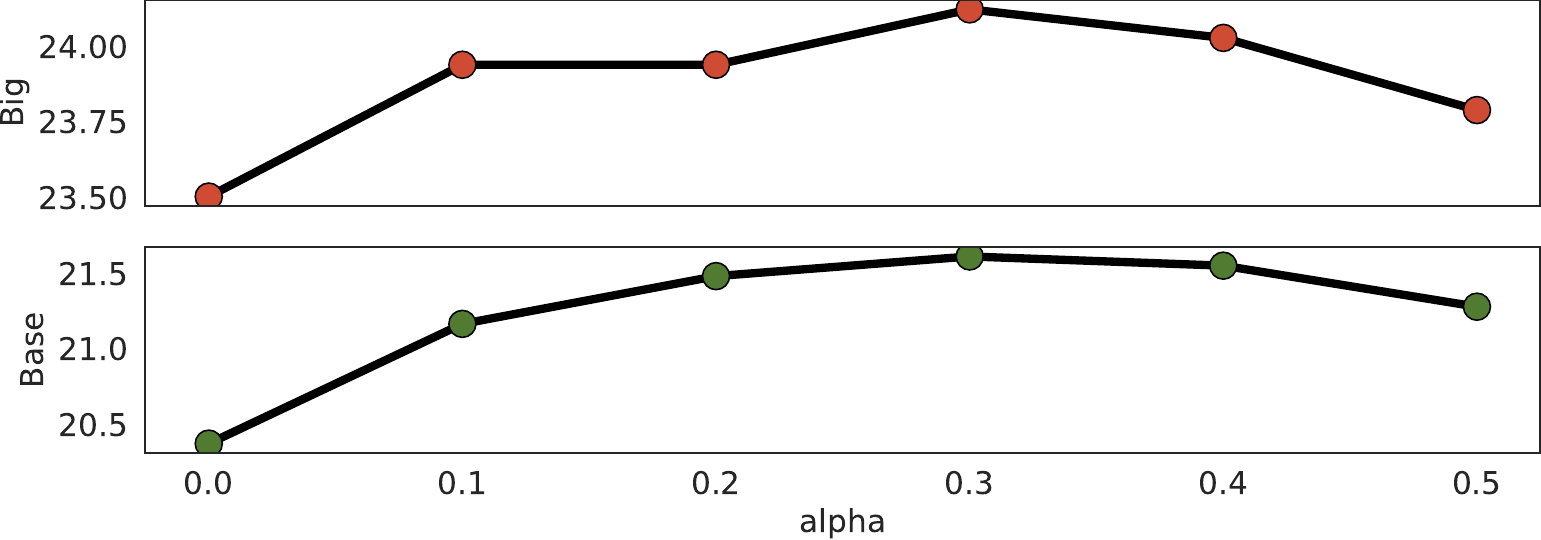}
  \label{fig:alpha-wmt}
  }
\caption{BLEU score along with the increase of pruning rate $\alpha$. Large $\alpha$ indicates small sub-network. Small dataset requires a larger $\alpha$ to yield better performance. IWSLT uses Transformer-small and WMT uses Transformer-base and Transformer-big.}
\label{fig:alpha}
\end{figure}

To better understand the pruning rate, we plot the performance along with the increase of pruning rate in Figure~\ref{fig:alpha}. 
For WMT, the best choice for $\alpha$ is 0.3 for both Transformer-base and Transformer-big, while for IWSLT the best $\alpha$ lies between 0.6$\sim$0.7. 
The results are consistent with our intuition, that large scale training data need a smaller pruning rate to keep the model capacity. 
Therefore, we suggest tuning $\alpha$ based on both the dataset and model size.  For large datasets such as WMT, setting a smaller $\alpha$ is better, while a larger $\alpha$ will slightly decrease the performance  (i.e. less than 0.5 BLEU score).
For small datasets like IWSLT, setting a larger $\alpha$ may yield better performance. 

\section{Conclusion}

In this paper, we propose to learn Language-Specific Sub-network (\method) for multilingual NMT.
%which automatically learns a language specific sub-network for each language pair.
Extensive experiments on IWSLT and WMT have shown that \method is able to alleviate parameter interference and boost performance. 
Further, \method can generalize well to new language pairs by training with a few hundred steps, while keeping the performance  of existing language pairs. Surprisingly, in zero-shot translation, \method surpasses the multilingual baseline by up to 26.5 BLEU.
Extensive analytic experiments are conducted to understand the characteristics of language specific sub-network.
Future work includes designing a more dedicated end-to-end training strategy and incorporating the insight we gain from analysis to design a further improved \method.

\bibliographystyle{acl_natbib}
\bibliography{acl2021}

\newpage
\appendix
\section{Appendices}

\subsection{Datasets Details}
\newcommand{\tabtl}[1]{\begin{tabular}[h]{@{}l@{}} #1 \end{tabular}}
\newcommand{\MC}{\multicolumn}
\newcommand{\MR}{\multirow}

\begin{table}[h]
    % \footnotesize
    % \scriptsize
    \setlength{\tabcolsep}{3.5pt}
    \centering
    \begin{tabular}{l l l l l}
    \toprule 
         \bf ISO & \bf Language & \bf  Family & \bf  Script & \bf  Size \\
    \midrule 
 fa & Farsi & Iranian & Arabic  & 89k  \\
 ar & Arabic & Arabic & Arabic  & 140k  \\
 he & Hebrew & Semitic & Hebrew  & 144k  \\
   \cmidrule{1-5} \\[-10pt]
 nl & Dutch & Germanic & Latin  & 153k  \\
 de & German & Germanic & Latin  &  160k \\
 \cmidrule{1-5} \\[-10pt]
  it & Italian & Romance & Latin  & 167k  \\
 es & Spanish & Romance & Latin  & 169k  \\
  \cmidrule{1-5} \\[-10pt]
 pl & Polish & Slavic & Latin  & 128k  \\

    \bottomrule 
    \end{tabular}
    \caption{Statistics and Language Family of IWSLT. Languages grouped together are similar languages.}
    \label{tab:iwslt-data}
\end{table}

\begin{table*}[!ht]
    % \footnotesize
    % \scriptsize
    \centering
    \begin{tabular}{l l l l l l l c}
    \toprule 
 \bf ISO & \bf Language & \bf  Family & \bf  Script & \bf Train &  \bf Valid & \bf Test & \bf  Size \\
    \midrule 
 gu & Gujarati & Indo-Aryan & Gujarati & WMT19 & newsdev19 & newstest19 & 11k  \\
 ta & Tamil & Dravidian & Tamil & WMT20 & newsdev20 & newstest20 & 64k  \\
\midrule

 kk & Kazakh & Turkic &	Cyrillic & WMT19  & newsdev19 &	newstest19 & 120k  \\
 tr & Turkish & Turkic &	Latin & WMT16 & newsdev16 &	newstest16 & 205k  \\
  \midrule

 ro & Romanian & Romance & Latin & WMT16  & newsdev16 &	newstest16 & 597k \\
  es & Spanish & Romance & Latin & WMT13  & newstest12 &	newstest13 & 13m  \\
 fr & French & Romance & Latin & WMT14  & newstest13 &	newstest14 & 37m  \\
  \midrule
  
 ps & Pashto & Iranian & Arabic & WMT20 & newsdev20	& newstest20 & 1m  \\
 \midrule

 fi & Farsi & Uralic &	Latin & WMT16 & newstest15 &	newstest16 & 2m  \\
 lv & Latvian & Baltic &	Latin & WMT17 & newsdev17 &	newstest17 & 2m  \\
 et & Estonian &  Uralic &	Latin & WMT18 & newsdev18	& newstest18 & 2.1m  \\
   \midrule
 lt & Lithuanian & Baltic &	Latin & WMT19  & newsdev19 &	newstest19 & 2.3m  \\
 
  \midrule

 ru & Russian & Slavic &	Cyrillic & WMT16 & newstest15 &	newstest16 & 2.5m \\
 \midrule
  cs & Czech & Slavic &	Latin & WMT14 & newstest13 &	newstest14 & 11m  \\
 pl & Polish & Slavic &	Latin & WMT20 & newsdev20 &	newstest20 & 11.1m  \\
  \midrule
  
 ja & Japanese & Japonic &	Kanji; Kana & WMT20 & newsdev20	& newstest20 & 16.8m  \\
  zh & Chinese & Chinese &	Chinese & WMT17 & newsdev17	& newstest17 & 20.8m  \\
  
\midrule

 de & German & Germanic &	Latin & WMT16 & newstest13 &	newstest14 & 4.5m  \\

    \bottomrule 
    \end{tabular}
    \caption{Statistics and Language Family of WMT daatset. Languages grouped together are similar languages.}
    \label{tab:wmt-data}
\end{table*}

% \usepackage{multirow}
\begin{table}[]
\renewcommand{\arraystretch}{1.0}
\setlength{\tabcolsep}{4pt}
\begin{tabular}{cc|cccccc}
\toprule
\multicolumn{1}{l}{}              & \multicolumn{1}{l}{}    & \multicolumn{6}{c}{Tgt}                          \\
                                  & \multicolumn{1}{c|}{}   & Fr & Cs         & De         & Es         & Ru         & Zh   \\ \midrule
\multirow{6}{*}{\rotatebox[origin=c]{90}{Src}} & \multicolumn{1}{c|}{Fr} &    & nt13 & nt13 & nt13 & nt13 & opus \\ 
                                  & \multicolumn{1}{c|}{Cs} &    &            & nt13 & nt13 & nt13 & ted  \\ 
                                  & \multicolumn{1}{c|}{De} &    &            &            & nt13 & nt13 & opus \\ 
                                  & \multicolumn{1}{c|}{Es} &    &            &            &            & nt13 & ted  \\ 
                                  & \multicolumn{1}{c|}{Ru} &    &            &            &            &            & opus \\ 
                                  & \multicolumn{1}{c|}{Zh} &    &            &            &            &            &      \\ \bottomrule
\end{tabular}
\caption{Datasets used in Zero-shot Translation. ``nt13'' indicates newstest2013.}
\label{tab:zero-shot-dataset}
\end{table}

\subsection{Training Details}

As stated in the previous section, we first train a multilingual baseline (Phase 1). Then we fine-tune the baseline on specific language pair to obtain the mask (Phase 2). After that we train the LaSS model with the obtained masks (Phase 3). Note that we only apply masks on linear weights, which means that the embedding weights, layer normalization are not masked out. We also exclude the output projection weight. We apply label smoothing of value 0.1 in all our experiments.

\subsubsection{IWSLT}
\paragraph{Model} We adopt Transformer-small \footnote{Transformer-base with $d_{ff}=1024$ and $n_{head}=4$} with dropout 0.1.
\paragraph{Data} Following \citet{tan2019multilingual}, we first tokenize the data then apply BPE. The BPE vocab size is 30k. We apply over-sampling with a temperature of $T=2$.
\paragraph{Training} For Phase 1, we train the baseline with Adam with a learning rate schedule of (5e-4,4k). The max tokens per batch is set to 262144.  For Phase 2, we keep all other settings unchanged except we set the max tokens to be 16384 and the dropout 0.3. For Phase 3, we keep the same setting as Phase 1, except we apply masks on the model.

\subsubsection{WMT}
\paragraph{Model} We adopt Transformer-base and Transformer-big with pre-norm \cite{DBLP:conf/acl/WangLXZLWC19}. We replace fixed positional embedding with learnable one and replace ReLU with GeLU. Also we use Layernorm-embedding \cite{DBLP:journals/corr/abs-2001-08210} to stabilize training.

\paragraph{Data} We use SentencePiece \cite{DBLP:conf/emnlp/KudoR18} to preprocess the data and learn BPE. Since the WMT dataset is highly imbalanced, we apply a temperature-based sampling strategy with $T=5$. To ensure all languages are represented adequately in the vocabulary, we apply the same temperature-based sampling strategy for training the BPE model.

\paragraph{Training} 
For Phase 1, we train the baseline with Adam with a learning rate schedule of (5e-4,8k). The max tokens per batch is set to 524288. For Phase 2, the warm-up updates are set to 1000.
To guarantee that the model does not overfit the data, we train on different language pairs with different steps and different batch size. Concretely, we fine-tune on $>$10k, $>$100k, $>$1m, $>$10m language pairs with 1k, 2k, 4k, 8k steps and max tokens per batch with 20480, 40960, 81920 and 163840.
For Phase 3, we keep the setting the same as Phase 1.

\subsection{Case Study}

\begin{CJK}{UTF8}{gkai}
\begin{table*}[!ht]

% \rowcolors{2}{lightgray}{}
\begin{center}
\begin{tabular}{lp{13cm}}
\toprule

\multicolumn{2}{c}{\textbf{Fr $\rightarrow$ Zh}} \\

\midrule

Src & La production annuelle d'acier était le symbole incontesté de la vigueur économique des nations. \\
Ref & 钢的年产量是国家经济实力的重要象征 \\
Baseline & Annual steel production was the undisputed symbol of nations' economic strength. \\
\method & 年度钢铁生产是各国经济活力的无可争辩的象征. \\

\midrule

Src & De l'avis de ma délégation donc, l'ONU devrait élargir ces activités de la façon suivante. \\
Ref & 因此,我国代表团认为,联合国现在应该以下述方式扩大这些活动。 \\
Baseline & 因此, in my delegation's view, the United Nations should expand these activities in the following manner. \\
\method & 因此,我国代表团认为,联合国应该扩大这些活动,如下. \\

\midrule

Src & Le domicile de la femme dépendait du lieu du mariage et de la résidence familiale. \\
Ref & 妇女的住处取决于婚姻和家庭位置。 \\
Baseline & The woman's place of residence depended on the place of marriage and family residence. \\
\method & 妻子的住所取决于婚姻地点和家庭住所. \\

\midrule
\midrule

\multicolumn{2}{c}{\textbf{De $\rightarrow$ Zh}} \\

\midrule

Src & Du bist gebissen worden. \\
Ref & 你被咬了 \\
Baseline & You have been bitten. \\
\method & 你被咬了 \\

\midrule

Src & Einmal würde schon reichen.  \\
Ref & 你只需要道歉一次就够了! \\
Baseline & Once upon a time it would be enough. \\
\method & 一次就足够了. \\

\midrule

Src & Wenn wir warten, hat er Zeit zum Tanken und Munitionieren. \\
Ref & 如果我们等待,他就有了 时间加油和补给弹药 \\
Baseline & When we wait, he has time for tanks and ammunition. \\
\method & 当我们等待时,他有时间去坦克和弹药. \\

\midrule
\midrule

\multicolumn{2}{c}{\textbf{Ru $\rightarrow$ Zh}} \\

\midrule

Src & Помощник заместителя министра здравоохранения Саудовской Аравии Его Превосходительство д-р Якуб бен Юсуф аль-Масрува \\
Ref & 沙特阿拉伯卫生部助理副部长雅各布·本·优素福·马斯如瓦博士阁下 \\
Baseline & Dr Yakub bin Yusuf al-Masruva, Deputy Minister of Health of Saudi Arabia \\
\method & 沙特阿拉伯卫生部副部长的助理,His Excellency Dr Yakub bin Yusuf al-Masruva \\

\midrule

Src & Не хочу я, чтобы Пит показывал нам фото, Элли. \\
Ref & 我不要皮特给我们看照片 艾莉 \\
Baseline & I don't want Pete showing us a photo, Elly. \\
\method & 我不想让皮特给我们看一下照片,艾丽. \\

\midrule

Src & Роджерс! Я сказал, встать в строй! \\
Ref & 罗杰斯 我说跟上 \\
Baseline & 罗吉尔斯! I said, get up! \\
\method & 罗吉尔斯,我说,你要站起来! \\

% Src &  \\
% Ref &  \\
% Baseline &  \\
% \method &  \\

\bottomrule
\end{tabular}
\caption{Case Study}
\label{tab:more-case-study}
\end{center}
\end{table*}
\end{CJK}

\end{document}